\documentclass[10pt,twocolumn,letterpaper]{article}

\usepackage{cvpr}
\usepackage{times}
\usepackage{epsfig}
\usepackage{graphicx}
\usepackage{amsmath}
\usepackage{amssymb}

\usepackage{multirow}
\usepackage{subcaption}
\usepackage{flushend}

\newcommand{\todo}[2][]{%
  \ifthenelse{\equal{#1}{}}
    {{\color{red}TODO:~#2}}
    {{\color{red}TODO~[#1]:~#2}}
}

\usepackage[pagebackref=true,breaklinks=true,letterpaper=true,colorlinks,bookmarks=false]{hyperref}

\cvprfinalcopy % *** Uncomment this line for the final submission

\def\cvprPaperID{} % *** Enter the CVPR Paper ID here

\ifcvprfinal\fi
\begin{document}

%%%%%%%%% TITLE
\title{Training Object Detectors on Synthetic Images\\ Containing Reflecting Materials}

\author{Sebastian Hartwig\\
\and
Timo Ropinski\\
}

\maketitle
%\thispagestyle{empty}

%%%%%%%%% ABSTRACT
\begin{abstract}
One of the grand challenges of deep learning is the requirement to obtain large labeled training data sets. While synthesized data sets can be used to overcome this challenge, it is important that these data sets close the reality gap, i.e., a model trained on synthetic image data is able to generalize to real images. Whereas, the reality gap can be considered bridged in several application scenarios, training on synthesized images containing reflecting materials requires further research. Since the appearance of objects with reflecting materials is dominated by the surrounding environment, this interaction needs to be considered during training data generation. Therefore, within this paper we examine the effect of reflecting materials in the context of synthetic image generation for training object detectors. We investigate the influence of rendering approach used for image synthesis, the effect of domain randomization, as well as the amount of used training data. To be able to compare our results to the state-of-the-art, we focus on indoor scenes as they have been investigated extensively. Within this scenario, bathroom furniture is a natural choice for objects with reflecting materials, for which we report our findings on real and synthetic testing data.
\end{abstract}

%%%%%%%%% BODY TEXT
\section{Introduction}
Training and validation of deep convolutional neural networks (CNNs) typically require a huge amount of labeled data. While for some domains, acquiring such amounts of labeled image data might not be a hurdle, in other domains this requirement poses one of the grand challenges of modern deep learning. Depending on the task, data labeling complexity varies in accordance with the required annotation, \eg class name, bounding box, segmentation mask, object pose, or semantic annotation. Labeling by hand is a time consuming process and can sometimes only be done by experts. Additionally, data labeling needs to be done carefully in order to achieve optimal results.

One effective way to overcome the lack of labeled data is to train on synthesized images. Exploiting image synthesis for data generation has two major benefits. First, the amount of image data can be controlled, and is only limited by synthesis performance and data storage. Second, the generated data is automatically labeled, as all synthesis parameters are known. Consequently, researchers have investigated the benefits and downsides of using synthesized data for different tasks. One such task is object detection, on which we also focus within this paper. For instance, in the works of Tremblay \etal\cite{Tremblay2018}, Peng \etal\cite{Peng2015}, Tobin \etal\cite{Tobin2017} and Hinterstoisser \etal\cite{Hinterstoisser2017}, 3D models representing real-world objects were used to generate synthetic image data, in order to train leading object detectors like Faster-RCNN, SSD, or Yolo3.

However, in order to make training on synthesized image data effective, the gap between synthetic and real data needs to be closed. One particular use case, where this has not been achieved yet, is object detection of objects exhibiting specular materials~\cite{shih2015reflection}. In contrast to more diffuse materials, specular materials pose additional challenges, as specular objects reflect the environment, and thus cannot be detected independent of it. Thus, within this paper, we focus on object detection tasks for specular objects, whereby we investigate how this task can benefit from synthesized training data. Specifically, we focus on object detection within interior scenes, as this area has also been worked on by other researchers, which made available large training and benchmark data sets, such that we can compare our results to the state-of-the-art. When focusing on interior scenes, a natural selection of specular objects to consider, is bathroom furniture, as these objects are prevalent and usually exhibit a specular material, which reflects the environment. Additionally, since bathroom furniture usually exhibits rounded down shapes, such objects are more challenging to detect, and thus the obtained results can be considered transferable to other object classes.

To investigate the usage of synthesized imagery for training object detectors for bathroom furniture, we consider three different image synthesis strategies, which vary in complexity and output realism. We have used two non-physically correct image synthesis approaches using BRDFs as material functions. One combines BRDFs with environment maps, which are frequently used to synthesize reflections, and the other one instead uses domain randomization~\cite{Tobin2017}, to expose the model to a wide range of environments at training time. As a third approach, we investigate a physically-correct Monte Carlo ray-tracing approach, which we have also combined with domain randomization. To investigate the descriptive power of the models trained with our synthesized image data, we have altered the domain randomization parameters, the number of classes to detect, as well as the number of training samples. We further compare the descriptive power of models trained with our training data, to models trained with readily available indoor scene training data sets.

Within the remainder of this paper, we will first discuss the work related to our approach in Section~\ref{sec:relatedwork}, before we detail the varieties of synthetic training data, which we have generated in Section~\ref{sec:synthetic-data-generation}. The object detection task results as obtained with the synthesized training data are reported for real and synthetic test data in Section~\ref{sec:experiments}. Finally, the paper concludes in Section~\ref{sec:conclusions}.

%-------------------------------------------------------------------------
\section{Related Work}\label{sec:relatedwork}
Within this section, we will first outline previous work focusing on the detection and reconstruction of specular objects. Then we will describe work related to the generation of synthetic image data for the purpose of training neural networks. Finally, we will discuss domain randomization approaches, as they have proven helpful in minimizing the amount of required training data sets.

\noindent\textbf{Specular object detection.} Reflecting materials result in the fact that objects strongly vary in visual appearance depending on their environment. With increasing specular value the effect of reflecting the surrounding world increases. Mirrors pose an extreme case where a perfect mirror shows a perfect reflection of the world. Whelan \etal\cite{whelan2018reconstructing} describe mirrors as "[..] essentially 'invisible'.[..]". Glass constitutes another special case of reflecting material causing detection problems due to its refraction property. Shih \etal\cite{shih2015reflection} proposed a method for reflection removal exploiting a reflection layer and a transmission layer of photographs taken through a window. Their method is also demonstrated on synthetic images. Removal or suppression is a common way to deal with reflections \cite{arvanitopoulos2017single}, \cite{xue2015computational}. Failing to detect highly reflecting surfaces poses heavy problems also in the field of 3D scene reconstruction. \textit{ScanNet}~\cite{dai2017scannet} a data set of annotated 3D reconstruction of indoor scenes shows many example scans with artifacts due to reflection. Whelan \etal\cite{whelan2018reconstructing} use \textit{AprilTags} in order to detect mirrors or glass surfaces to successfully reconstruct highly reflecting surfaces and the world they are located in.

\noindent\textbf{Synthetic data generation.} In recent years, the amount of work focusing on training object detectors on synthetic data has increased. While some work focused on optical flow prediction, exploiting synthetic data for end-to-end training of a CNN \cite{butler2012naturalistic}, \cite{dosovitskiy2015flownet}, \cite{mayer2016large}, others are using synthetic data for understanding indoor scenes \cite{handa2016understanding}, \cite{mccormac2016scenenet}. In all these approaches, image synthesis algorithms with different degrees of sophistication are used to generate the required training data. So often data sets acquired through physically-based rendering are introduced~\cite{tremblay2018deep},~\cite{zhang2016physically}. Ray tracing the scene for instance yields realistic looking training images. Zhang \etal\cite{zhang2016physically} use pure physically-based renderings and Tremblay \etal\cite{Tremblay2018} uses domain randomization and physically-based images to bridge the reality gap. In contrast to Tremblay \etal\cite{tremblay2018deep} we randomize a physically-based simulator. That enables us to apply domain randomization to photorealistic images.

While simulators of physical correct lighting produce realistic looking images, the generation of these data sets is time and resource consuming. Therefore, often dedicated hardware is exploited in order to ensure reasonable compute times ahead of training. Besides rendering, the scene complexity and material parameters are also crucial to obtain realistic looking results. To refine synthetic images in order to make them look more realistic, Nogues \etal\cite{Nogues2018} propose to use a generative adversarial network. Existing data sets like Virtual KITTI~\cite{gaidon2016virtual}, Falling Things~\cite{tremblay2018falling}, \cite{to2018ndds}, and many more~\cite{qiu2016unrealcv}, \cite{dosovitskiy2015flownet}, \cite{zhang2016unrealstereo} have been proven to be valuable, as object detector models trained on these data sets perform well on real imagery. By using these data sets, researchers have further investigated the possibility of fine-tuning pre-trained models. In the work of Hinterstoisser \etal\cite{Hinterstoisser2017} weights of a feature extractor pre-trained on real images were frozen in order to train the remaining object detector on synthetic images only. There results indicate that fine-tuning the feature extractor by unfreezing it, degrades the performance of the detector significantly.

\noindent\textbf{Domain randomization.} To improve the synthesis of image data for training purposes, a method called domain randomization was introduced by Tobin \etal\cite{Tobin2017}, that got picked up by Tremblay \etal\cite{Tremblay2018}. The basic idea is to randomize the synthesizer to expose the model to a wide range of environments at training time. Then, during validation on real images the model generalizes to real images and interprets them as another variation of trained synthetic images. Thus, Tobin \etal\cite{Tobin2017} could address the reality gap in robotic learning performed on simulated data, where they could show that an object detector trained on synthetic images only, achieved also high accuracy on real images. Tremblay \etal\cite{Tremblay2018} applied domain randomization in order to close the reality gap between real images and synthetic images in the KITTY data set. The performance of an object detector trained on the virtual KITTI data set is compared to one that was trained on domain randomized images. Testing on the KITTI data set shows that domain randomization is able to bridge the reality gap. Tremblay \etal\cite{tremblay2018deep} also exploit domain randomization to explore the reality gap in the context of 6-DoF pose estimation. Therefore, they combined randomized with photorealistic images in order to train a pose estimator achieving state-of-the-art performance on 6-DoF object pose estimation.

%-------------------------------------------------------------------------
\section{Synthetic Data Generation}\label{sec:synthetic-data-generation}
\begin{figure}[t]
\begin{center}
\includegraphics[width=\linewidth]{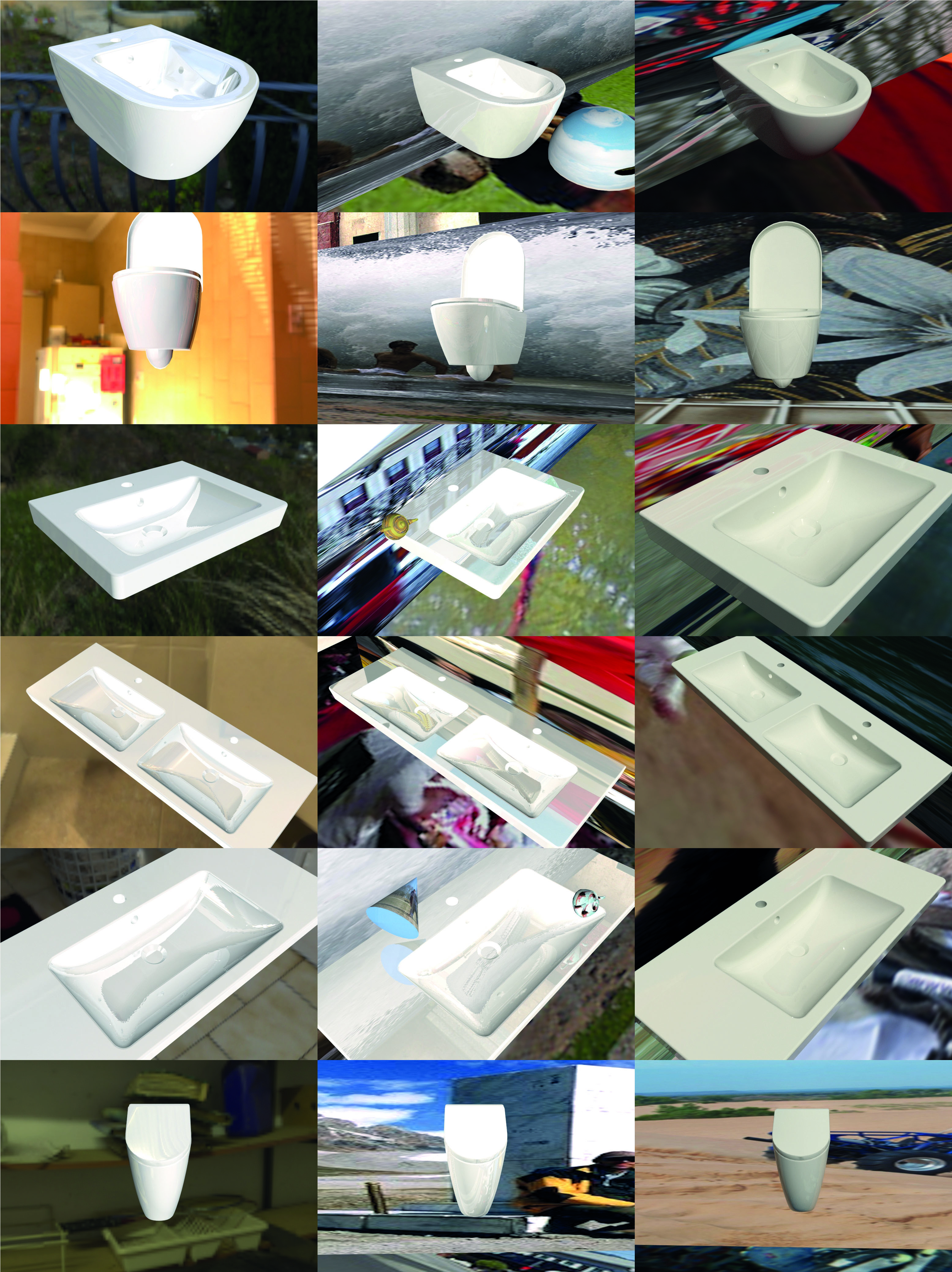}
\end{center}
   \caption{Example images synthesized with our three image synthesis protocols (\textit{columns}) for each of the 6 models used in the first study (\textit{rows}). First column (\textbf{RA}) shows images generated using BRDFs and environment mapping only, while the second column adds scene geometry and domain randomization (\textbf{DR}). The third column shows the image resulting from additionally using a photorealistic renderer (\textbf{MLT-DR}).}
\label{fig:data sets}
\end{figure}
The examination of render methods for reflecting materials yield insights about when synthetic data fails and when it can be successfully used in deep learning approaches. Thus, to investigate the best parameters for synthesizing object detector training data for reflecting materials, we make use of a state-of-the-art Monte Carlo ray-caster, which enables us to generate large amounts of photorealistic images. Independent of the degree of realism of the individual images, synthetic images often suffer from insufficient or missing details when simulating the real world. In order to counteract this effect, which is also denoted as the reality gap, domain randomization was introduced~\cite{Tobin2017}. Accordingly, we combine our different image synthesis protocols with domain randomization in order to investigate the synthesis process. Thus, we have devised three different image synthesis approaches by combining rendering and domain randomization. In the following sections we describe how we generate the image data used during our investigation. While we differentiate three ways to synthesize training images, all three techniques assume the following data collection steps to be done in advance.

\noindent\textbf{Data collection.} Among indoor scenes, bathrooms typically exhibit the largest concentration of reflecting materials, such as porcelain, ceramic, chrome, or even glass. Thus, to be able to benchmark with existing data sets, bathroom furniture was a natural choice to investigate object detection of reflecting objects. To synthesize realistic bathroom scenes, we have carefully chosen CAD models from the sanitary area, which we have obtained from manufacturer homepages. We selected $6$ models from the popular \textit{Villeroy \& Boch - Subway 2.0 collection} for the pre-study and our first experiments. Later, we added more models from other common manufacturers \textit{Duravit, Hansgrohe and Grohe} which add up to $21$ models as shown in Figure \ref{fig:intra-classes}. To ensure a common scale, models were re-scaled, and sometimes if necessary smoothed. Material properties contained in the model descriptions were discarded and instead predefined materials were used as described below. As for each data set we wanted to consider models which have a high probability to be present in a sanitary scene, we have chosen the \textit{6} distinct classes shown in Figure~\ref{fig:data sets} for our first experiment.

\subsection{Protocol 1: Reflection Approximation (RA)}\label{ssec:reflection-approximation}
When synthesizing training data sets, one obvious goal is to keep computation times low, as this can have a great impact when synthesizing the large amounts of data needed for training. Consequently, our first protocol is a rather simplistic reflection approximation by means of local shading in combination with environment mapping. The environment mapping is realized by applying sphere mapping, as it is often done in computer graphics applications. Since reflections in this approach are not based on the actual scene geometry, but solely the environment map, we have not considered any other scene geometry besides the actual bathroom furniture model. Thus, to synthesize the training images, the individual models are centered in the origin of the scene. During render time different rendering parameters are chosen to be altered randomly per frame. We place two point lights in a static position in front of the model. Intensity values per light source are set randomly in a range between $0.0$ and $1.0$. Since bathroom furniture is usually mounted to walls we did not consider views from the back of the object. While the camera is always oriented towards the center of the model, the position of the camera was uniformly sampled from within a hemispherical volume directed to the front side of the model. For shading we used a local Blinn-Phong reflection model~\cite{blinn1977models} combined with BRDF sampling. To make the background consistent with the reflections, we used the environment maps also for the background coloring. Textures for BRDF and environment mapping are taken from the \textit{Flickr8K} data set~\cite{hodosh2013framing} which consists of $8K$ images. We use $250$ camera positions per model. Then, for each of the $6$ CAD models around $1.5K$ frames are generated. We decide to use the aspect ratio of our real images for synthesis. We also wanted some of the images to be quadratic as it is common used input ratio for CNNs. Images are rendered both in landscape and portrait using the following dimensions: $(518,346)$,$(300,300)$ or $(493,326)$ pixels. The first column in Figure \ref{fig:data sets} shows synthesized example images for each of the 6 models.

% \usepackage{graphicx}

\subsection{Protocol 2: Domain Randomization (DR)}
\begin{figure}[t]
\begin{center}
\includegraphics[width=\linewidth]{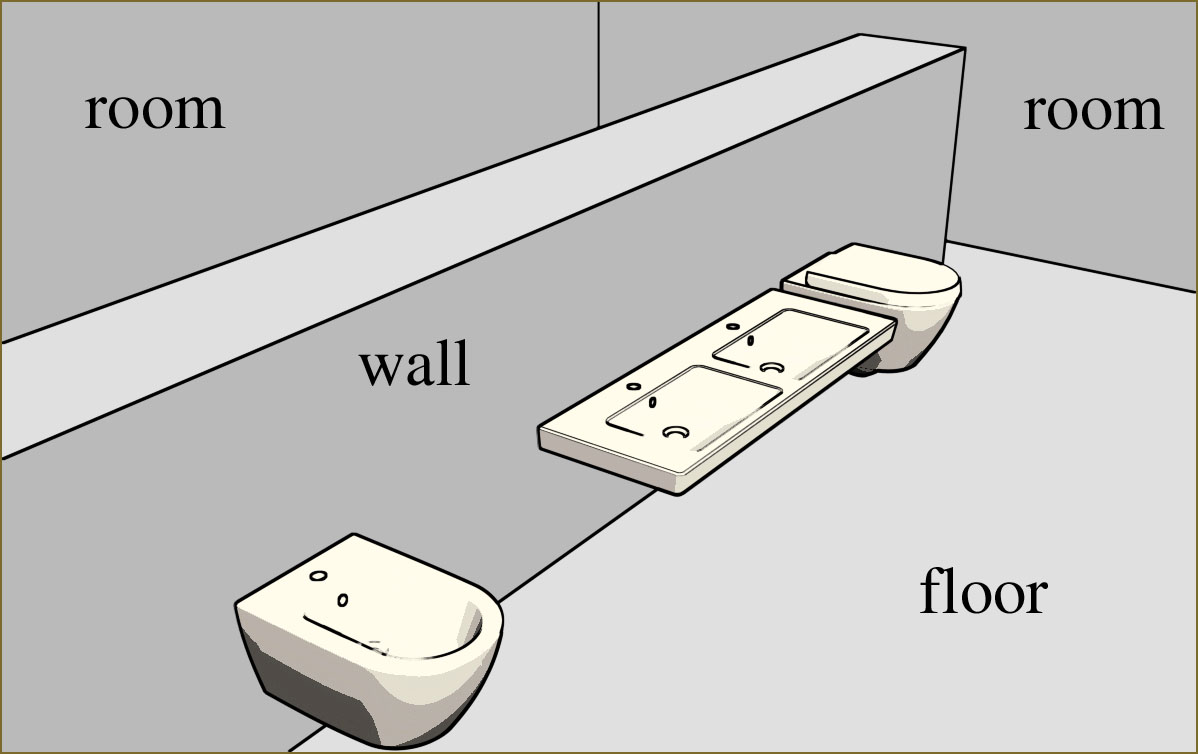}
\end{center}
   \caption{Our virtual bathroom scene consists of a cube with an additional wall inside. During render time CAD models of our 6 object classes are randomly placed onto that wall. Additionally a random number of occluders of different type and size are thrown into the scene. For each generated frame textures for room, wall floor and occluders are chosen randomly from a large set of images. The resulting data set contains around $38K$ randomized images.}
\label{fig:dr-scene}
\end{figure}
Domain randomization was introduced to bridge the gap between synthetic and real images~\cite{Tobin2017}. Thus, in our second protocol we apply domain randomization in order to train an object detector that can differentiate strong resembling object classes. Domain randomization is used to achieve heavy variability at training time, such that at test time the model generalizes to real images. We also considered a slightly more complex scene then before, such that we can obtain another degree of parameterization. Our setup consists of a room, a wall and a floor plane as shown in Figure \ref{fig:dr-scene}. For each render pass we shuffle position of the objects mounted to the wall. Therefore a random number $n \in [1..n_{c}]$ of models is chosen, where $n_c=6$ is the number of classes. Then $n$ models are placed randomly along the full width of the wall whereby we prevent models to overlap. Additionally, we select a random number $m \in [5..20]$ of occluders which are placed in front of the wall plane. We use the following types of occluders: pyramid, box, cone, cylinder, sphere, teapot, torus and tube, whereby we alter the scaling for each instance. Again, textures are randomly sampled from the \textit{Flickr8K} data set~\cite{hodosh2013framing}. To prevent overfitting we triple the data set size used in Section \ref{ssec:reflection-approximation}. For each generated frame we alter randomly the following rendering parameters:
\begin{itemize}
  \itemsep0em
  \item Material color of the object classes. One of $75$ different shades of white, see Figure \ref{fig:color-palette}.
  \item Reflection value between $0.0$ and $1.0$.
  \item Camera position and look vector which are sampled from within the volume in front of the \textit{wall}.
  \item Camera roll angle $\pm30$ degree.
  \item Texture of the floor, wall, room and occluders.
  \item Frame dimensions: $\left(518,346\right),\left(300,300\right) or \left(493,326\right)$ pixels in portrait or landscape.
\end{itemize}

\subsection{Protocol 3: Physically-Based Rendering with Domain Randomization (MLT-DR)}
\label{sec:physical-reflection}
In this protocol, we wanted to investigate how much can be gained by employing a photo-realistic renderer. For this purpose we used the same simple bathroom scene as shown in Figure~\ref{fig:dr-scene}. Additionally, spot lights and physical material properties like specularity, reflectivity, and metalness values are used for randomization. Thus during image synthesis, we altered rendering parameters as described before, and additionally changed the following parameters:
\begin{itemize}
  \itemsep0em
  \item Number of light sources ranging from $3$ to $13$.
  \item Position of light sources, not lower than model positions.
  \item Direction of light sources, targeting towards a random spot on the wall.
  \item Reflection, metalness and specular value of physical material (used for models only).
\end{itemize}

\begin{figure}
    \centering
    \includegraphics[width=\linewidth]{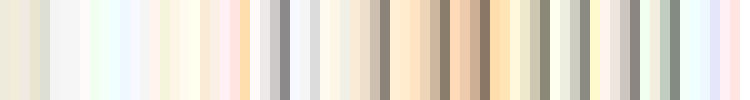}
    \caption{Color palette which is used to simulate ceramic material. In total we used $75$ shades of white, gray and beige.}
    \label{fig:color-palette}
\end{figure}

While randomizing occluders, wall, floor, camera, lighting and positioning, colorization of our models is done in a domain specific way in order to maintain the photo-realistic look of ceramic. Therefore, material color of the models are selected from an appropriate color palette of different shapes of white, gray and beige, see Figure \ref{fig:color-palette}. The color tones have been selected based on our experience, we collected when taking and processing over 1000 photographs of real-world bathroom furniture. With the given tones, we could replicate the visual appearance of these images the best.

\subsection{Sub-Class Challenge (SC)}
\begin{figure*}
\begin{center}
\includegraphics[width=0.7\linewidth]{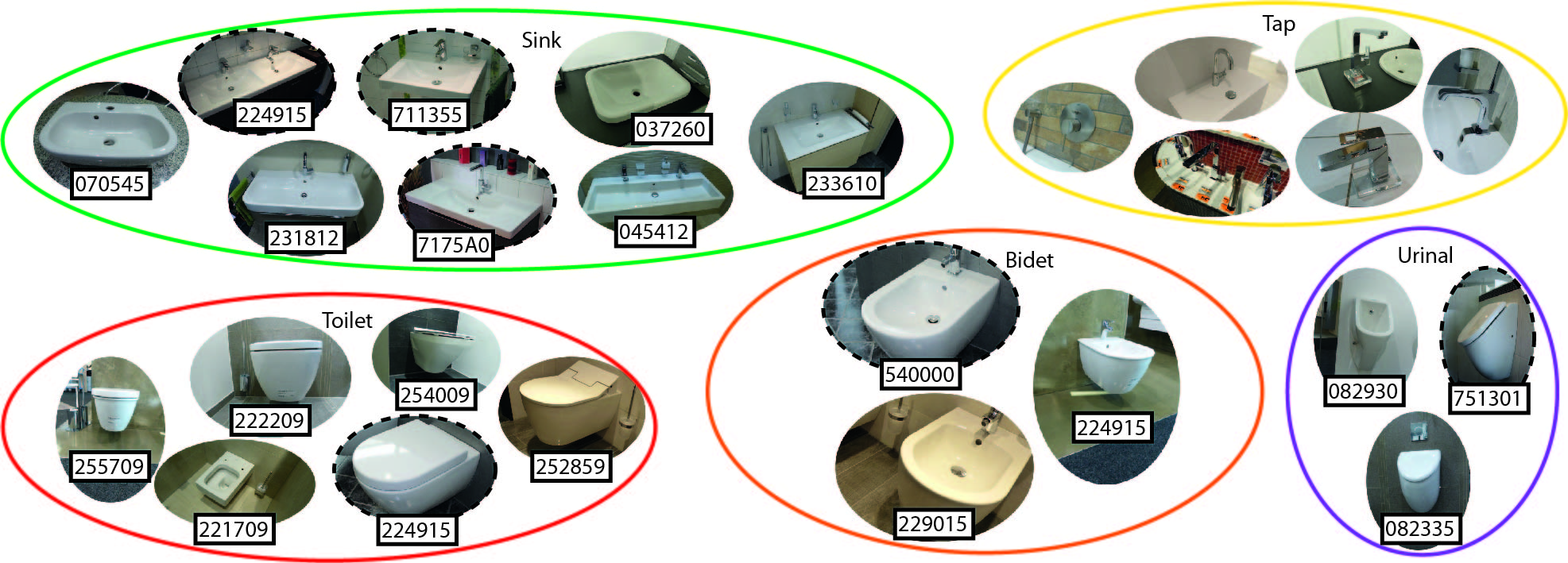}
\end{center}
   \caption{In total we have tested the described image synthesis approaches on 21 different classes. The ellipses show, how we have clustered the classes for our external validity study, where we have compared against less finely labeled training data.}
\label{fig:intra-classes}
\end{figure*}

While we used only 6 classes in our first experiment, in order to find optimal synthesis parameters, we have also extended the range of classes and thus added more complexity to the detection task. Therefore, we distinguish between 5 classes which are divided into 21 sub-classes as shown in Figure~\ref{fig:intra-classes}. Therefore, we obtained further CAD models from manufacturer homepages and applied the same data preparation steps as described in the beginning of this section. In order to test out the limits of our approach we chose models from two different manufacturers only. Since bathroom installations can visual easily be differentiated by the way how they are installed (e.g. wall-mounted, detached, integrated, etc.), we select CAD models that feature the same wall-mounted version preventing single models to stick out. We select for class \textit{sink} $8$ models resulting in $8$ sub-classes divided into: $3$ small sinks, $3$ large sinks and $2$ double sinks. For class \textit{toilet} we chose $6$ models split into $2$ cornered toilets and $4$ rounded. Class \textit{urinal} consists of $3$ versions which are separated in $2$ having a lid and $1$ without lid. Finally, class \textit{bidet} we assign $3$ similar versions. For an overview of our classes, see Figure \ref{fig:intra-classes}. Since we have also included a class labeled \textit{Tap} we provide suitable material, like brass, stainless steel and chrome. In order to examine performance on models with an increased metalness value, we chose $8$ different models for class \textit{Tap} and merged them to one class. For rendering we use the method described in \textbf{MLT-DR} whereby we kept all parameters untouched, except for the number of object classes. The generated data set consists of roughly $100K$ images.

%-------------------------------------------------------------------------
\section{Experiments}\label{sec:experiments}
Our experiments are divided into two studies: a pre-study where we have investigated the influence of reflective materials, and the main study, where we have exploited different image syntehsis techniques for these materials.

\subsection{Classification Pre-study}
To investigate the role of reflective materials in the context of deep learning, we have conducted a pre-study, in which we we initially focused on training a feature extractor on synthetic images. For the feature extraction we used \textit{InceptionV3}, which is pre-trained on \textit{ImageNet}. For each run we use a \textit{categorical cross entropy} loss function and an \textit{AdaDelta} optimizer with $learning rate=1.0$, $rho=0.95$, $epsilon=0.0000001$ and $decay=0.0$. We further set $batch size=32$, $steps=15625$ and image dimensions to $dim = (200,200)$. We want to examine prediction accuracy depending on camera position, selected background color and method for computing reflection. Randomizing position of the camera is done by sampling a point on the surface of a front-facing hemisphere around the object. Additionally, \textit{radius} of the hemisphere is set to a value reported in Table~\ref{tab:feature-extractor}. Column \textit{reflection} holds the value to decide whether reflection is simulated through BRDF (TRUE), diffuse material only (FALSE) or both (MIXED). Next column \textit{background} determines if the background is just black (BLACK), random solid color (COLOR), environment mapping (ENVMAP) or either of them (COLOR + ENVMAP). Textures for BRDF sampling and environment mapping are also taken from the \textit{Flick8K} data set~\cite{Silberman:ECCV12}. We have also analyzed the frame aspect ratio of training images (\textit{aspect}), as well as the field of view of the camera, which is set to $fov \in {45.0, 60.0, 63.0}$ degrees. Classes and their corresponding CAD models from Section \ref{sec:synthetic-data-generation} are used for synthesis. Each feature extractor is then fine tuned on a set of synthetic images, for a specific number of training images, see fifth column of Table~\ref{tab:feature-extractor}. For validation we used synthetic and real images, whereby real images come in two versions: patches and original images. A patch is a down-scaled version of a real image with size of a training image ($200$x$200$ pixels) and the object centered in the middle. During training accuracy on train images (column \textit{acc}) and validation images (column \textit{val\_acc}) is specified in Table~\ref{tab:feature-extractor}. For validation on real images mean average precision is reported for patches and original images in second last and last column of Table \ref{tab:feature-extractor}.

\noindent While the feature extractor has picture-book performance during training time on synthetic images, it struggles to classify correctly on real images. Nevertheless, an increase in performance is to be observed examining the last two columns. All in all we could identify increasing performance of the extractor when randomization is increases. When surveying the upper half of Table~\ref{tab:feature-extractor}, it becomes clear that reflection is a crucial parameter in terms of learning from synthetic data, which confirms our approach presented in Section~\ref{sec:synthetic-data-generation}.

\begin{table*}[]
\resizebox{\textwidth}{!}{%
\begin{tabular}{|c|c|c|c|c|c|c|c|c|c|}
\hline
radius & reflection & background & aspect & \# train & \# validation & acc & val\_acc & mAP patches & mAP full \\ \hline
{[}1.0, 1.5, 2.0, 2.5, 3.0, 3.5, 4.0{]} & MIXED & COLOR + ENVMAP & 1.0 & 5000 & 1000 & 0.9870 & 0.9234 & \textbf{0.3218} & \textbf{0.3655} \\ \hline
{[}1.0, 1.5, 2.0, 2.5, 3.0, 3.5, 4.0{]} & TRUE & COLOR + ENVMAP & 1.5 & 5000 & 1000 & 0.9679 & 0.9435 & 0.3821 & 0.3549 \\ \hline
{[}1.0, 1.5, 2.0, 2.5, 3.0, 3.5, 4.0{]} & TRUE & COLOR + ENVMAP & 1.0 & 5000 & 1000 & 0.9844 & 0.9536 & \textbf{0.4051} & \textbf{0.3398} \\ \hline
{[}1.0, 1.5, 2.0, 2.5, 3.0, 3.5, 4.0{]} & TRUE & COLOR + ENVMAP & {[}1.0, 1.5{]} & 10000 & 2000 & 0.9774 & 0.9441 & 0.3563 & 0.2901 \\ \hline
{[}2.0{]} & MIXED & ENVMAP & 1.0 & 5000 & 1000 & 0.9968 & 0.9788 & 0.0958 & 0.2582 \\ \hline
{[}1.0, 1.5, 2.0, 2.5, 3.0, 3.5, 4.0{]} & MIXED & ENVMAP & 1.0 & 5000 & 1000 & 0.9890 & 0.9486 & 0.1277 & 0.2537 \\ \hline
{[}1.0{]} & FALSE & BLACK & 1.0 & 5000 & 1000 & 0.9980 & 0.9970 & 0.1916 & 0.2058 \\ \hline
{[}2.0{]} & MIXED & COLOR + ENVMAP & 1.0 & 5000 & 1000 & 0.9940 & 0.9768 & 0.0761 & 0.2040 \\ \hline
{[}1.0{]} & MIXED & COLOR + ENVMAP & 1.0 & 5000 & 1000 & 0.9944 & 0.9808 & 0.1302 & 0.1818 \\ \hline
{[}1.0, 1.5, 2.0, 2.5, 3.0, 3.5, 4.0{]} & FALSE & COLOR + ENVMAP & 1.0 & 5000 & 1000 & 0.9972 & 0.9929 & 0.1474 & 0.1774 \\ \hline
{[}1.0, 1.5, 2.0, 2.5, 3.0, 3.5, 4.0{]} & FALSE & ENVMAP & 1.0 & 5000 & 1000 & 0.9966 & 0.9919 & 0.1498 & 0.1641 \\ \hline
{[}1.0{]} & MIXED & ENVMAP & 1.0 & 5000 & 1000 & 0.9910 & 0.9879 & 0.0933 & 0.1579 \\ \hline
{[}2.0{]} & FALSE & BLACK & 1.0 & 5000 & 1000 & 1.0 & 1.0 & 0.1425 & 0.1552 \\ \hline
{[}1.0, 1.5, 2.0, 2.5, 3.0, 3.5, 4.0{]} & FALSE & COLOR & 1.0 & 5000 & 1000 & 0.9980 & 0.9950 & 0.1891 & 0.1543 \\ \hline
{[}2.0{]} & FALSE & ENVMAP & 1.0 & 5000 & 1000 & 0.9984 & 0.9980 & 0.1130 & 0.1535 \\ \hline
{[}1.0{]} & FALSE & ENVMAP & 1.0 & 5000 & 1000 & 0.9962 & 0.9970 & 0.0737 & 0.1384 \\ \hline
{[}2.0{]} & FALSE & COLOR & 1.0 & 5000 & 1000 & 0.9976 & 1.0 & 0.1646 & 0.1215 \\ \hline
{[}1.0{]} & FALSE & COLOR & 1.0 & 5000 & 1000 & 0.9990 & 0.9980 & 0.1449 & 0.1197 \\ \hline
{[}1.0, 1.5, 2.0, 2.5, 3.0, 3.5, 4.0{]} & FALSE & BLACK & 1.0 & 5000 & 1000 & 0.9976 & 1.0 & 0.1425 & 0.0878 \\ \hline
{[}2.0{]} & FALSE & COLOR + ENVMAP & 1.0 & 5000 & 1000 & 0.9980 & 0.9929 & 0.1425 & 0.0834 \\ \hline
{[}1.0{]} & FALSE & COLOR + ENVMAP & 1.0 & 5000 & 1000 & 0.9986 & 0.9919 & 0.0884 & 0.0727 \\ \hline
\end{tabular}%
}
\caption{To gauge the influence of reflective materials, we examined randomization of image synthesis within a pre-study. Therefore, we trained an \textit{InceptionV3} feature extractor pre-trained on \textit{ImageNet} using different rendering parameters. Per row we altered rendering parameters and report accuracy of the feature extractor during training on synthetic images (\textit{val\_acc}) and the mean average precision on real images (\textit{mAP patch, mAP full}).}
\label{tab:feature-extractor}
\end{table*}

%
%Parameter Study
%- Photorealism: BRDF+EnvMaps, Local Shaging+DomRan, Arnold Renderer+DomRan (Physically-Correct, Materials and Spotlights) (for 6 classes)
%    - Domain randomization: number of occluders, textures, lighting, camera positions
%- Number of classes: 6 vs. 21 classes (only for Arnold Renderer+DomRan)
%- Origin Trainingsdata: Princeton DS (2 classes), MIT(Limits of data variability -> international)
%- Amount of training images: 20k (87\%), 75k, 100k (90\%)
%
\begin{figure}
\begin{center}
\includegraphics[width=0.8\linewidth]{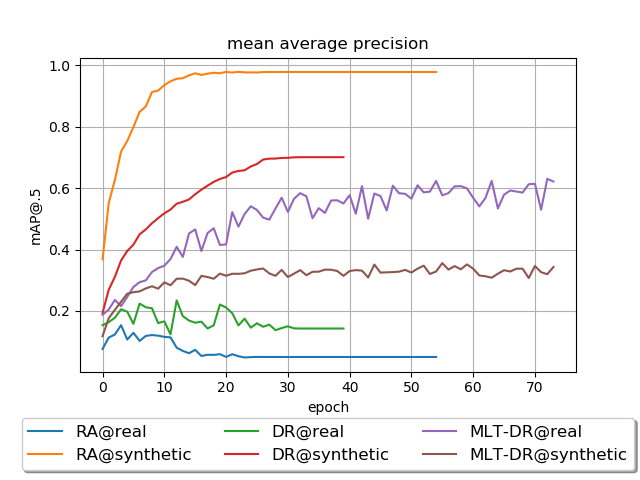}
\end{center}
   \caption{We report performance of the \textit{RetinaNet} object detector on our three data sets: \textbf{(RA)}, \textbf{(DR)} and \textbf{(MLT-DR)} for synthetic images as well as for real images. Performance is measured in mean average precision with $IoU >= 0.5$, \textit{mAP@.5}. Training of each object detector is terminated when training has stopped improving.}
\label{fig:precision}
\end{figure}

\subsection{Object Detection Study}
Motivated by the knowledge obtained from the pre-study described above, we focus on object detection in our main experiment. In our renderings we detected that in many cases the background class dominates. To over come this imbalance we trained the one-stage object detector \textit{RetinaNet}~\cite{lin2018focal} on each our three data sets: \textit{RA}, \textit{DR} and \textit{MLT-DR}. For validation on real images we took over $1000$ photographs in exhibitions of local distributors and labeled them manually. Figure~\ref{fig:intra-classes} shows such example photographs of our chosen models, while Figure~\ref{fig:data sets} shows synthesized versions of the models which are marked with a dashed box in Figure \ref{fig:intra-classes}.

\subsubsection{Training Parameters}
Within this section we describe the training parameters used to train the \textit{RetinaNet} on our three synthesized training data sets \textit{RA}, \textit{DR} and \textit{MLT-DR}.

\noindent\textbf{RA.} Six classes shown in Figure~\ref{fig:data sets} are used for this experiment on training images synthesized using local BRDFs with environment maps for reflection approximation. In order to train the object detector we used $12K$ frames. We set $batch\_size = 8$, from which we determine the number of steps per epoch $steps = 1500$. Training is stopped when no further improvement takes place. For \textit{RA} the process was stopped after $55$ epochs.

\noindent\textbf{DR.} In the second experiment we trained another detector on the \textit{DR} images. A simple indoor scene is randomized in order to apply domain randomization to our data set as described in Section \ref{sec:synthetic-data-generation}. For synthesis of frames we also used local shading. We train the object detector with $38K$ images and we set $batch\_size = 8$ leading to $4750$ steps per epoch. This task was terminated after $39$ epochs.

\noindent\textbf{MLT-DR.} Finally, in our third experiment we trained a third object detector on the \textit{MLT-DL} images for $73$ epochs with the same $batch\_size$ and $step$ size as in the run before.

\begin{table}[]
\centering
\resizebox{0.7\linewidth}{!}{%
\begin{tabular}{c|c|c|c|}
\cline{2-4}
 & (RA) & (DR) & (MLT-DR) \\ \hline
\multicolumn{1}{|c|}{toilet (AP@.5)} & 0.40  & 0.34  & 0.85  \\ \hline
\multicolumn{1}{|c|}{bidet (AP@.5)} & 0.11  & 0.30  & 0.86  \\ \hline
\multicolumn{1}{|c|}{urinal (AP@.5)} & 0.06  & 0.01  & 0.38  \\ \hline
\multicolumn{1}{|c|}{double sink (AP@.5)} & 0.10  & 0.29  & 0.55  \\ \hline
\multicolumn{1}{|c|}{small sink (AP@.5)} & 0.03  & 0.08  & 0.57  \\ \hline
\multicolumn{1}{|c|}{large sink (AP@.5)} & 0.06  & 0.38  & 0.54  \\ \hline \hline
\multicolumn{1}{|c|}{mAP@.0} & 0.28  &  0.30  &  0.66  \\ \hline
\multicolumn{1}{|c|}{mAP@.25} & 0.20 &  0.26  &  0.65  \\ \hline
\multicolumn{1}{|c|}{mAP@.5} &  0.12  &   0.23  & \textbf{ 0.63 } \\ \hline
\multicolumn{1}{|c|}{mAP@.75} & 0.02 & 0.14 &  0.59\\ \hline
\end{tabular}%
}
\caption{The first six rows of this table show results of testing the object detector on real images. For each class the average precision with $IoU \in {0.0, 0.25, 0.5, 0.75}$ is reported. In the last four rows we display mean average precision on all three data sets with $IoU \in {0.0, 0.25, 0.5, 0.75}$.}
\label{tab:map-6-class}
\end{table}

\subsubsection{Results}
Details about performance on our real image data set are displayed in Figure~\ref{fig:precision} for all three object detectors. For the sake of completeness we report \textit{average precision (AP)} and \textit{mean average precision (mAP)} from the best pass in each training process in Table~\ref{tab:map-6-class}. For box regression we used the common \textit{smooth L1 loss} function. Categorical loss is computed as described in~\cite{lin2018focal}, see Equation~\ref{eq:focal_loss}, where the \textit{focus} parameter is set to $\gamma = 2.0$ and weighting factor $alpha = 0.25$.

\begin{equation}
    FL\left ( p_{t} \right )= -\alpha_{t}\left ( 1-p_{t} \right )^{\gamma }log\left ( p_{t} \right )
    \label{eq:focal_loss}
\end{equation}

Examining Figure~\ref{fig:precision} it is clear that there is still a gap between synthetic and real images. While in our \textit{RA} data set discrepancy between synthetic and real images is almost at maximum, our \textit{DR} data set narrows the gap but leaving performance of real images beyond synthetic images. In contrast, \textit{MLT-DR} shows better results on real images than on synthetic ones, closing the gap even further.

%\begin{figure*}
%\includegraphics[width=0.33\textwidth]{images/ra-loss.png}
%\includegraphics[width=0.33\textwidth]{images/dr-loss.png}
%\includegraphics[width=0.33\textwidth]{images/mlt-dr-loss.png}
%\caption{Comparison of training and validation loss during training of the object detector. In order to train the detector we use %a focal loss for classification and smooth L1 for box regression. In the left Figure loss is plotted for \textbf{RA}. The middle %plot shows loss for \textbf{DR} and the right plot shows loss for \textbf{MLT-DR}.} \label{fig:losses}
%\end{figure*}

\subsection{Sub-Class Challenge}
\begin{figure*}
\begin{center}
\includegraphics[width=\linewidth]{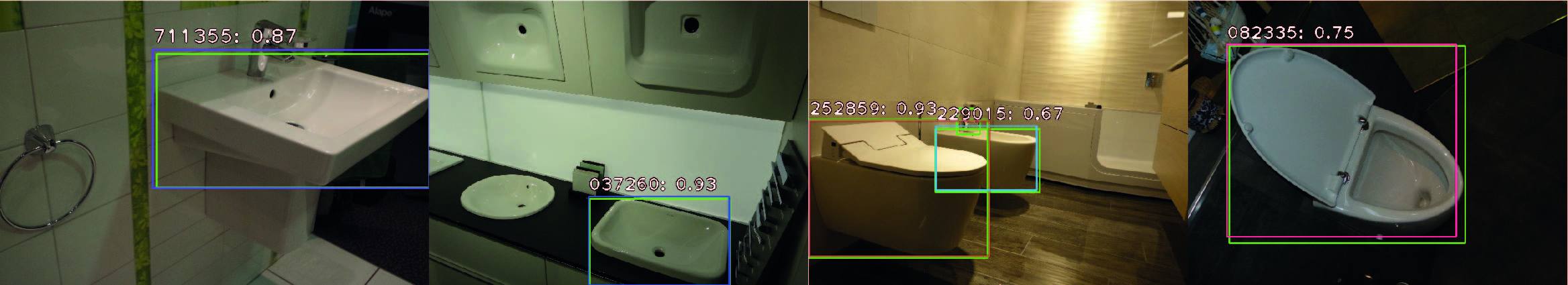}
\end{center}
   \caption{In this Figure we show some detection results of our sub-class challenge. We display ground truth in green and for the prediction we report the predicted class name and score.}
\label{fig:detections}
\end{figure*}

As described in Section~\ref{sec:synthetic-data-generation}, we also prepared a data set including $21$ classes. We trained two object detectors on this \textit{SC} data set. For the first detector we used $20$ classes ignoring the \textit{tap} class annotations and in a second run we trained another object detector on all $21$ classes in order to analyze the effect of having highly reflecting materials present. For an overview of the distribution of our sub-classes we refer to Figure~\ref{fig:intra-classes}. For training of both detectors roughly $100K$ synthetic images are used. In detail, the first detector is trained using a data set containing renderings synthesized from models belonging to classes \textit{sink, toilet, urinal and bidet}. Each class is divided in up to $8$ sub-classes. Our second detector is trained on the same data set. However, we additionally included the \textit{tap} class to the set of classes, such that we have $21$ classes in total. Here we regard this class only as a single class, and refer to the supplementary material for further details on this sub-class challenge, where we report the achieved performance on more than $21$ classes featuring heavy reflecting materials like \textit{stainless steel and chrome}. We plot performance of both detectors in Table \ref{tab:sub-class-results} and show some detections on real images in Figure \ref{fig:detections}. Summarizing the results, we can say both detectors are able to detect sub-classes in our real-images, but still there is some investigations to be done regarding heavily reflecting materials like stainless steel and chrome.

% \usepackage{graphicx}
\begin{table*}[]
\centering
\resizebox{\textwidth}{!}{%
\begin{tabular}{c|c|c|c|c|c|c|c|c|c|c|c|c|c|c|c|c|c|c|c|c|c|l}
\cline{2-22}
\multicolumn{1}{l|}{} & \multicolumn{6}{c|}{Toilet} & \multicolumn{3}{c|}{Bidet} & \multicolumn{3}{c|}{Urinal} & \multicolumn{8}{c|}{Sink} & Tap &  \\ \cline{2-23}
 & 5614R0 & 222209 & 255709 & 221709 & 254009 & 252859 & 540000 & 224915 & 229015 & 082335 & 082930 & 751301 & 711355 & 7175A0 & 7175D0 & 233610 & 231812 & 070545 & 037260 & 045412 & 000000 & \multicolumn{1}{c|}{mAP@.5} \\ \hline
\multicolumn{1}{|c|}{AP@.5} & 0.69 & 0.4 & 0.3 & 0.41 & 0.43 & 0.82 & 0.61 & 0.32 & 0.10 & 0.71 & 0.41 & 0.53 & 0.83 & 0.58 & 0.71 & 0.61 & 0.86 & 0.82 & 0.98 & 0.84 & - & \multicolumn{1}{l|}{0.60} \\ \hline
\multicolumn{1}{|c|}{AP@.5} & 0.68 & 0.33 & 0.36 & 0.01 & 0.22 & 0.73 & 0.59 & 0.35 & 0.06 & 0.69 & 0.40 & 0.47 & 0.61 & 0.41 & 0.58 & 0.25 & 0.73 & 0.58 & 0.97 & 0.64 & 0.14 & \multicolumn{1}{l|}{0.47} \\ \hline
\end{tabular}%
}
\caption{In this Table we report the performance on sub-class challenge. Two object detectors were trained on $20$ and $21$ classes in order to locate and classify bathroom objects in real images.}
\label{tab:sub-class-results}
\end{table*}

\begin{figure}
\begin{center}
\includegraphics[width=0.8\linewidth]{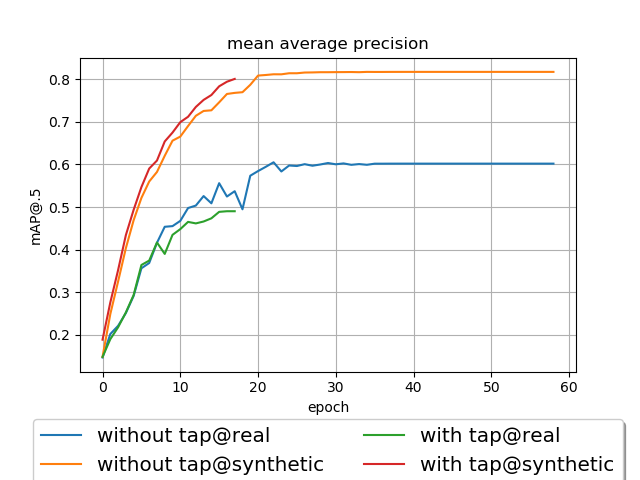}
\end{center}
   \caption{In this Figure we report performance difference between two detectors. One trained on $20$ and the other on $21$ sub-classes.}
\label{fig:precision-20}
\end{figure}

\subsection{External Validation}

\begin{figure}
\begin{center}
\includegraphics[width=0.8\linewidth]{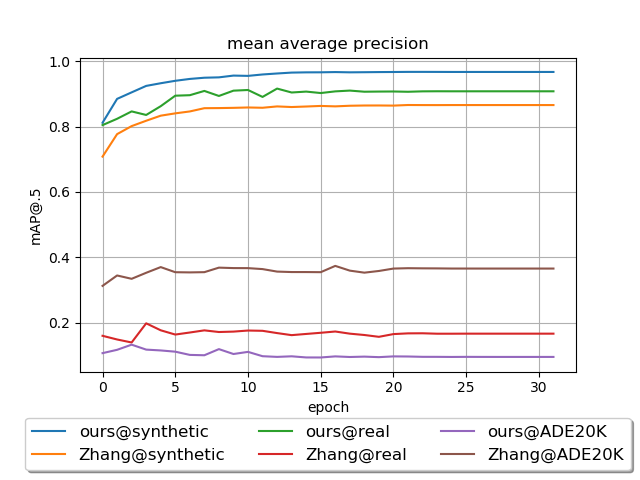}
\end{center}
   \caption{We train two object detector one on our \textit{SC} data set and the other on the data set of Zhang \etal\cite{zhang2016physically}. In this Figure we report performance of both detectors on synthetic images denoted as \textit{synthetic}, real images of our data set denoted as \textit{real} and real images of \textit{ADE20K}\cite{zhou2017scene} denoted as \textit{ADE20K}.}
\label{fig:precision-ours-theirs}
\end{figure}

To further validate our procedural synthesis of physically-based renderings for object detector training, we compare the \textit{SC} data set with Zhang \etal's~\cite{zhang2016physically} physically-based data set. They sampled roughly $500K$ images from $45K$ realistic indoor scenes, varying render methods, and lighting conditions. Their data set consists of $40$ classes (e.g. \textit{Wall, Floor, Cabinet, Bed, etc.}) according to the NYUv2 data set~\cite{Silberman:ECCV12}. Since we focused on reflecting materials, and our data set consists of models from the sanitary area, we compared our \textit{SC} class list with the class list of Silberman\etal\cite{Silberman:ECCV12}, and ended up with an intersection of $2$ classes: \textit{Toilet} and \textit{Sink}. Filtering out corresponding images from Zhang \etal's data set~\cite{zhang2016physically} yields roughly $75K$ images containing at least one of the before mentioned labels. While we use our data set from our sub-class challenge, we randomly discard images in order to match the number of images in Zhang \etal's data set~\cite{zhang2016physically}. We then remap class labels like following: \textit{sink} $\longrightarrow$ \textit{sink}, \textit{toilet} $\longrightarrow$ \textit{toilet}, \textit{urinal} $\longrightarrow$ \textit{toilet}, \textit{bidet} $\longrightarrow$ \textit{toilet}, since their data set already includes \textit{toilet} and \textit{sink} labels. Having both data sets prepared we train two object detectors in the same way as described in the experiments before. Setting $batch\_size = 8$, and using \textit{focal loss function}~\cite{lin2018focal} for classification and \textit{smooth L1} for box regression. We trained both detectors for approximately $270K$ steps after which they did not further improve. For validation we generated around $5K$ synthetic images. We compared performance on synthetic images with real images of ours and real images of the \textit{ADE20K}~\cite{zhou2017scene} data set from which we filtered out bathroom images containing the labels: \textit{toilet} and \textit{sink} which are around $700$ images. Performance results for each detector are reported in Figure \ref{fig:precision-ours-theirs}. Since we have compared both data sets on two classes only we refer to the supplementary material for extended comparison. It is clear that the detector trained on our data set performs well on our real images, but fails on images of the \textit{ADE20K} data set. The detector trained on Zhang \etal's~\cite{zhang2016physically} physically-based data set performs better than ours on \textit{ADE20K}, but fails on our real images. We argue due to our focus on sub-class detection the detector trained on our images was not able to generalize. On the other hand it successfully accomplished our sub-class challenge.

%-------------------------------------------------------------------------
\section{Conclusions}\label{sec:conclusions}
Within this paper, we have investigated image synthesis techniques for generating object detector training data, which exhibits reflecting materials. To understand the influence of rendering technique and domain randomization, we have investigated three different image synthesis protocols. Our tests indicate, that only a combination of photo-realistic rendering and domain randomization has the potential to train robust object detectors on synthetic data, such that they can be successfully applied to real-world images. Further, we have met the challenge of sub-class detection. We have successfully trained a detector on synthetic images that is able to locate and distinguish strong resembling models in real images.

{\small
\bibliographystyle{ieee}
\bibliography{main}
}

\end{document}